\documentclass{article}
\usepackage{ijcai24}

\usepackage{times}
\usepackage{soul}
\usepackage{url}
\usepackage[hidelinks]{hyperref}
\usepackage[utf8]{inputenc}
\usepackage[small]{caption}
\usepackage{graphicx}
\usepackage{amsmath}
\usepackage{amsthm}
\usepackage{booktabs}
\usepackage{algorithm}
\usepackage{algorithmic}
\usepackage[switch]{lineno}

\usepackage{multicol}
\usepackage{multirow}

\title{Meta In-Context Learning Makes Large Language Models \\Better Zero and Few-Shot Relation Extractors}

\author{
    Anonymous submission
}

\author{
Guozheng Li$^1$
\and
Peng Wang$^{1,2}$\thanks{Corresponding author} \and
Jiajun Liu$^1$\and
Yikai Guo$^3$\and
Ke Ji$^1$\and
Ziyu Shang$^1$\And
Zijie Xu$^1$\\
\affiliations
$^1$School of Computer Science and Engineering, Southeast University\\
$^2$Key Laboratory of New Generation Artificial Intelligence Technology and Its \\
Interdisciplinary Applications (Southeast University), Ministry of Education\\
$^3$Beijing Institute of Computer Technology and Application\\
\emails
\{gzli, pwang, jiajliu, keji, ziyus1999, zijiexu\}@seu.edu.cn
}

\begin{document}

\maketitle

\begin{abstract}
    Relation extraction (RE) is an important task that aims to identify the relationships between entities in texts. While large language models (LLMs) have revealed remarkable in-context learning (ICL) capability for general zero and few-shot learning, recent studies indicate that current LLMs still struggle with zero and few-shot RE. Previous studies are mainly dedicated to design prompt formats and select good examples for improving ICL-based RE. Although both factors are vital for ICL, if one can fundamentally boost the ICL capability of LLMs in RE, the zero and few-shot RE performance via ICL would be significantly improved. To this end, we introduce \textsc{Micre} (\textbf{M}eta \textbf{I}n-\textbf{C}ontext learning of LLMs for \textbf{R}elation \textbf{E}xtraction), a new meta-training framework for zero and few-shot RE where an LLM is tuned to do ICL on a diverse collection of RE datasets (i.e., learning to learn in context for RE). Through meta-training, the model becomes more effectively to learn a new RE task in context by conditioning on a few training examples with no parameter updates or task-specific templates at inference time, enabling better zero and few-shot task generalization. We experiment \textsc{Micre} on various LLMs with different model scales and 12 public RE datasets, and then evaluate it on unseen RE benchmarks under zero and few-shot settings. \textsc{Micre} delivers comparable or superior performance compared to a range of baselines including supervised fine-tuning and typical in-context learning methods. We find that the gains are particular significant for larger model scales, and using a diverse set of the meta-training RE datasets is key to improvements. Empirically, we show that \textsc{Micre} can transfer the relation semantic knowledge via relation label name during inference on target RE datasets.
\end{abstract}

\section{Introduction}
Relation extraction (RE)~\cite{wang2020two,wei2020novel,li2022fastre,wang2023pascore} aims to identify the relationships between entities in texts, and plays an important role in natural language processing (NLP). Existing RE methods with pre-trained language models (PLMs)~\cite{devlin2019bert,liu2019roberta} have achieved outstanding performance by fully supervised fine-tuning. However, such a supervised paradigm heavily depends on large-scale annotated data. Hence, in real-world scenarios, existing methods tend to struggle when recognizing new relations with insufficient annotation resources (low-shot) which is coined as zero and few-shot RE~\cite{han-etal-2018-fewrel,chen-li-2021-zs}.

Two popular paradigm of methods are emerged to alleviate the challenge of zero and few-shot learning, including meta learning~\cite{finn2017model,snell2017prototypical} and in-context learning~\cite{brown2020language,wei2022chain}. 
Meta-learning provides a framework for learning to learn, which addresses the challenges of zero and few-shot learning by training models on multiple tasks with limited labeled data, enabling them to generalize to new, unseen classes or tasks. However, existing meta-learning-based RE studies~\cite{zhao-etal-2023-matching,zhang-etal-2023-hypernetwork} still typically focus on specific tasks, datasets and settings, lacking of flexibility and generalization to more general low-shot scenarios. Thereafter, in-context learning (ICL) which concatenates a query and few-shot demonstrations to prompt LLMs for prediction is proposed, where it is plug-and-play and does not require additional inductive bias learning or sophisticated template design. And recent studies~\cite{wei2022chain,kojima2022large} on large language models (LLMs), such as GPT-3~\cite{brown2020language} and ChatGPT~\cite{openai2022}, demonstrate that LLMs perform well in various downstream tasks without any training or fine-tuning but only with ICL. However, some recent researches~\cite{ma2023large,wang2023instructuie} have revealed a significant performance gap in LLMs when it comes to apply ICL to the low-shot RE tasks, while others~\cite{agrawal-etal-2022-large,li2023revisiting} believe that LLMs deliver promising performances in low-shot RE, because of the influence of query forms and selected demonstrations. But we argue that fundamentally improving the ICL capability of LLMs in RE is substantially important.

In this work, we borrow and combine the ideas of meta learning and in-context learning in RE, introduce a new low-shot RE framework with meta in-context learning~\cite{min2022metaicl,chen2022meta} called \textsc{Micre}: \textbf{M}eta \textbf{I}n-\textbf{C}ontext learning of LLMs for \textbf{R}elation \textbf{E}xtraction. Specifically, we reformulate the RE task as a natural language generation problem. Unlike previous approaches that fine-tune the models with task-specific augmentation, \textsc{Micre} tunes an LLM on a collection of RE datasets to learn how to in-context learning, and is evaluated on strictly new unseen RE datasets with zero and few-shot prompting~\cite{brown2020language,kojima2022large}. Each meta-training examples matches the inference setup where it includes several examples in training sets from one RE dataset that will be concatenated together as a single sequence to the LLM, and the output of the final example is used to calculate the cross-entropy training loss. Tuning the LLM in this manner directly leads to better ICL and low-shot RE, where the LLM learns to recover the semantics of the RE task from the given examples at inference time. This method is related to recent work~\cite{wei2022finetuned,sanh2021multitask} that uses multi-task learning for better zero-shot performance at inference time. However, \textsc{Micre} allows adapting to new RE datasets and domains from several examples alone under few-shot scenarios, without relying on a task reformatting (e.g., reducing the RE process to input-output format) or task-specific templates (e.g., converting the RE task to a natural language generation problem). For zero-shot cases, we induce the complete prediction results from LLMs through prompting known entities and relations, keeping consistent prompt formats with meta-training phases.

To unify the prompt formats of zero and few-shot RE, we design a simple tabular prompting~\cite{li2024unlocking} for extracting subjects, objects and predicates from texts. Specifically, a table header “$|$Predicate$|$Subject$|$Object$|$” is provided as part of the prompt and the LLMs automatically generate a table, where “$|$” is the recognizable delimiter of tables. This enables the effective zero-shot entity and relation extraction even without the participation of few-shot examples. We experiment \textsc{Micre} with tabular prompting on a collection of 12 publicly available RE datasets and evaluate it on unseen RE benchmarks to ensure no overlap between meta-training and target datasets. Experimental results show that \textsc{Micre} consistently outperforms baselines including ICL without meta-training and task-specific zero and few-shot RE with fine-tuning. This demonstrates \textsc{Micre} enables LLMs to recover the semantics of the entities and relations in context during inference. In summary, our contributions are three-fold: 
\begin{itemize}
    \item We introduce \textsc{Micre}, a new meta in-context training framework based on LLMs for zero and few-shot RE to learn how to do in-context learning, resulting in better low-shot prompting performance at new unseen RE tasks. We also devise a effective tabular prompting to unify the prompt formats of zero and few-shot RE.
    \item By meta in-context training \textsc{Micre} on various LLMs with different model scales and a collection of RE datasets, \textsc{Micre} delivers comparable or superior performance compared to state-of-the-art zero and few-shot RE baselines on unseen datasets.
    \item Further analysis shows that the gains are particular significant for larger model scales and diverse datasets. And \textsc{Micre} achieves good ICL results during inference due to its relation semantic knowledge transferring.
\end{itemize}

\section{Related Work}
\paragraph{Meta in-context learning.} Large language models (LLMs) perform well in various downstream tasks without any training or fine-tuning but only with a few examples as instructions, which is called in-context learning~\cite{brown2020language}. ICL has been further improved by later work and shows promising results on a variety of tasks~\cite{zhao2021calibrate,min2022metaicl,ji2023hierarchical,shang2024ontofact,liu2024towards}. However, ICL with LLMs achieves poor performance when the target task is very different from language modeling in nature such as entity and relation extraction. While prior work has shown that multi-task learning on a large collection of tasks leads to better performance on a new task when tested zero-shot~\cite{mishra2022cross,wei2022finetuned}, recent studies~\cite{min2022metaicl,chen2022meta} propose to explicitly train models on an in-context learning objective via multi-task learning, where this learning to learn in-context paradigm is typically called meta in-context learning. Our work is based on the core idea of meta in-context training regarding entity and relation extraction via multi-task learning, showing \textsc{Micre} achieves substantial improvements on zero and few-shot RE tasks.

\paragraph{Zero and few-shot relation extraction.} Few-shot relation extraction~\cite{han-etal-2018-fewrel,yu2020bridging,wang2023fmlre} aims to predict novel relations by exploring a few labeled examples. MAML~\cite{finn2017model} and prototypical networks~\cite{snell2017prototypical} are widely used and combined with pre-trained language models~\cite{devlin2019bert,liu2019roberta} in few-shot settings to achieve impressive results. To be capable of extracting relations that were not specified in advance, zero-shot relation extraction~\cite{levy-etal-2017-zero,chen-li-2021-zs,chia2022relationprompt} is proposed to invent new models to predict new relations. Besides the fine-tuned small language models for RE, recent studies~\cite{ma2023large,li2023revisiting,wan-etal-2023-gpt} leverage the LLMs with zero and few-shot prompting to extract entities and relations from texts through ICL. However, the current primary research of ICL for RE only consider two directions: query forms~\cite{li2023revisiting,li2024unlocking} and demonstration retrieval techniques~\cite{ma2023large,wan-etal-2023-gpt}. Although both factors are vital for achieving better ICL for RE performance, we argue that fundamentally improving the ICL capability of LLMs in RE task is substantially important. Therefore, we propose to train LLMs with in-context learning objective to improve the ICL ability of LLMs for better in-context RE. The main insight is to conduct a multi-task learning scheme over a collection of meta-training RE datasets, in order to learn how to condition on a small set of training examples, recover the semantics of a new RE task, and predict the output based on it.

\begin{figure*}[ht]
    \small
	\centering
	\includegraphics[width=0.99\linewidth]{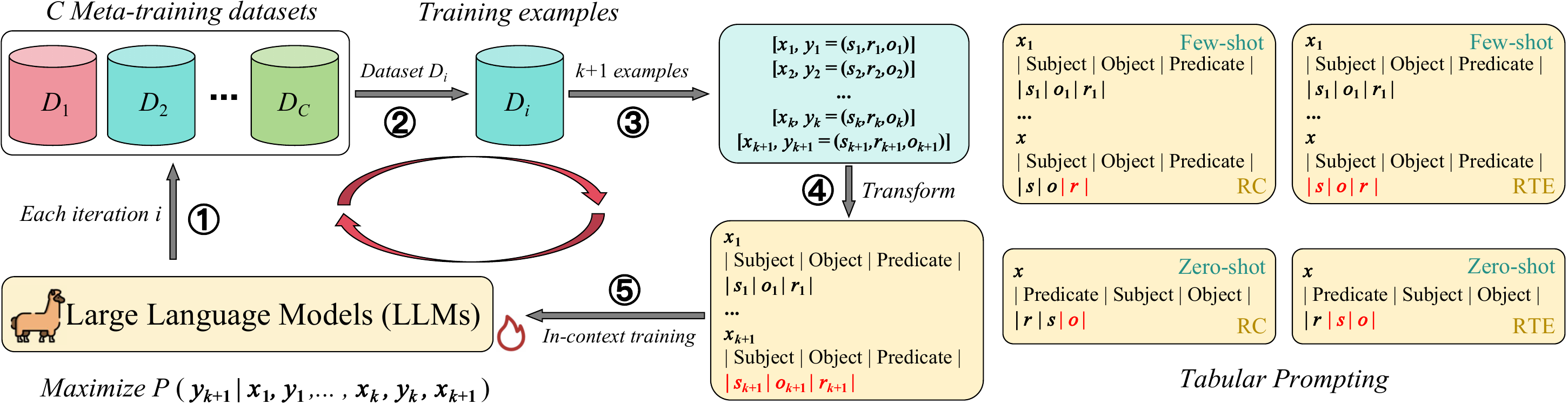}
	\caption{Overview of the meta-training work flow of \textsc{Micre}. The output of LLMs are highlighted in red.}
	\label{micre}
\end{figure*}

\section{Methodology}
\subsection{Task Formulations}
We consider two RE tasks: relation classification (RC) and relational triple extraction (RTE), under two low-shot settings: zero-shot setting and few-shot setting. (1) \textbf{Relation classification}: Given a sentence $S_i$ and an entity pair $(s_i, o_i)$ where $s_i$ is the subject and $o_i$ is the object. This task aims to identify a relation $r_i$ from pre-defined relation set $\mathcal{R}$ that satisfy the relationship between subject $s_i$ and object $o_i$ expressed by the sentence $S_i$. (2) \textbf{Relational triple extraction}: Given a sentence $S_i$, this task aims to jointly extract the relational triple $z_i = (s_i,r_i,o_i)$ from sentence $S_i$ where $s_i$ and $o_i$ are entities and $r_i$ is the corresponding relation from relation set $\mathcal{R}$. (3) \textbf{Low-shot settings}: For zero-shot setting, the model are expected to identify novel entities and relations at the inference time without any training examples. For few-shot setting, we formulate it in the typical $N$-way-$K$-shot form. Given the training set $\mathcal{D}$ and the target sentence set $\mathcal{T}$, for $N$ way (relation types), the model utilizes randomly $K$ examples for each relation type from $\mathcal{D}$ to form the support set $\mathcal{S}=\{(x_i, y_i)\}_{i=1}^{N \times K}$, where $x_i$ denotes the sentence $S_i$ and $y_i$ denotes the relation $r_i$ in RC or relational triple $z_i$ in RTE. The model are expected to utilize only $N \times K$ examples from $\mathcal{S}$ for prediction and outputs recognized relations in RC and triples in RTE for each target sentence in $\mathcal{T}$.

\subsection{Meta In-Context Training}
Following previous literature~\cite{brown2020language,min2022metaicl,chen2022meta}, the training examples are concatenated and provided as an single input to the model, which is feasible for $k$-shot learning. At test time, the model is evaluated on an unseen target RE task that comes with $N \times K$ training examples (i.e. few-shot learning) or no training examples (i.e. zero-shot learning), and inference directly follows the similar prompting format as in meta-training phases.

Figure~\ref{micre} provides an overview of the meta-training work flow of \textsc{Micre} which consists of five steps. The model is meta-trained on a collection of RE datasets which we call meta-training datasets. Specifically, for each meta-training iteration $i$ in step 1, we sample a dataset $D_i$ from $C$ meta-training datasets in step 2. Then $k + 1$ training examples $(x_1, y_1),... ,(x_{k+1}, y_{k+1})$ are sampled from the training examples of the chosen dataset in step 3. To unify different tasks (i.e. RC and RTE) and settings (i.e. zero and few-shot) for convenient transferring, we transform these $k+1$ samples into the tabular prompting format (introduce it later) in step 4. We then supervise the model (tune the LLMs) by feeding the concatenation of $x_1, y_1,... , x_k, y_k, x_{k+1}$ to the model as an input and train the model to generate $y_{k+1}$ using a negative log likelihood objective in the final step. 


\subsection{Tabular Prompting}
We adopt a tabular prompting for unifying two different tasks and settings that generates organized and concise outputs in ICL. Specifically, a table header “$|$Predicate$|$Subject$|$Object$|$” is provided to prompt the LLMs to automatically generate a table, where “$|$” is the recognizable delimiter of tables. This strategy is suitable for both zero and few-shot settings compared to solely text-to-text prompting format, as it provides precise instructed signals by table header for zero-shot prompting. During meta-training, we utilize two orders of table header “$|$Predicate$|$Subject$|$Object$|$” and “$|$Subject$|$Object$|$Predicate$|$” to improve the robustness of models and unify the RC and RTE tasks at inference time.

\subsection{Zero and Few-shot Inference}
For a new target RE task, the model is given $N \times K$ training examples $(x_1, y_1),... ,(x_{N \times K}, y_{N \times K})$ for few-shot prompting or no training examples for zero-shot prompting as well as a test input $x$. It is also given a set of candidates $\mathcal{C}$ which is either a set of relation labels (RC) or relational triples (RTE). For few-shot setting, as in meta-training, the model takes a concatenation of $x_1, y_1,... , x_{N \times K}, y_{N \times K}$, $x$ as the input, and compute the probability of each candidate $c_i \in \mathcal{C}$. The candidate with the maximum conditional probability is returned as a prediction. For zero-shot setting, the model only takes $x$ as the input with no training examples to get the prediction. 

The specific low-shot prompting forms in RC and RTE are illustrated in Figure~\ref{micre}. For few-shot prompting, the models easily recover the output of test input $x$ conditioning on $N \times K$ training examples. For zero-shot prompting, the situation becomes a little tricky. Since no training examples are available for in-context learning, the models are unware of relation schema in a specific RE dataset. We transform the multi classification form into multiple binary classification forms. Specifically, we prompt the models with each relation label $r$ to generate the corresponding subject $s$ and object $o$, then we select the correct relation $r$ or triple $z$ from $\mathcal{C}$ candidates. In RC task, given relation $r$ and subject $s$, we consider relation $r$ as the zero-shot prediction if its probability of output object $o$ of the original annotation (for RC, we know both annotated subject and object) is the maximum. In RTE task, we prompt the models with each relation $r$ to generate multiple candidate relational triples, then the triple $z$ with the maximum conditional probability is selected as the zero-shot prediction.


\section{Experiments}
\subsection{Experimental Design}
\paragraph{Datasets.} We use a collection of publicly RE datasets taken from~\cite{wang2023instructuie} and widely considered in RE research community. We have 12 unique RE datasets in total, covering general, news, disease and science domains. 
All these RE datasets are in English and we provide the statistics of these datasets in Table~\ref{data}. Note that in meta-training phases, we only use the training set of 12 RE datasets. To balance the meta-training datasets, we sample 10,000 examples for each training set and include all examples for training sets with fewer than 10,000 samples~\cite{wang2023instructuie}.

\begin{table}[h]
\small
\centering\setlength{\tabcolsep}{1.5mm}
\begin{tabular}{l|lll}
\toprule
{Dataset} & {\#Train} & {\#Dev} & {\#Test} \\
\midrule
{ADE~\cite{gurulingappa2012development}} & {3,417} & {427} & {428} \\
{CoNLL2004~\cite{roth2004linear}} & {922} & {231} & {288} \\
{GIDS~\cite{jat2018improving}} & {8,526} & {1,417} & {4,307} \\
{KBP37~\cite{zhang2015relation}} & {15,917} & {1,724} & {3,405} \\
{NYT24~\cite{riedel2010modeling}} & {56,196} & {5,000} & {5,000} \\
{NYT11~\cite{takanobu2019hierarchical}} & {62,648} & {149} & {369} \\
{SciERC~\cite{luan2018multi}} & {1,366} & {187} & {397} \\
{SemEval~\cite{hendrickx2019semeval}} & {6,507} & {1,493} & {2,717} \\
{TACRED~\cite{zhang2017position}} & {68,124} & {22,631} & {15,509} \\
{ACE2004~\cite{doddington2004automatic}} & {6,946} & {868} & {868} \\
{ACE2005~\cite{walker2005ace}} & {10,051} & {2,424} & {2,050} \\
{WebNLG~\cite{gardent2017webnlg}} & {5,019} & {500} & {703} \\
\bottomrule
\end{tabular}
\caption{Statistics of meta-training datasets.}
\label{data}
\end{table}

We experiment on FewRel~\cite{han-etal-2018-fewrel} and Wiki-ZSL~\cite{chen-li-2021-zs} for low-shot experiments. To ensure no overlap between meta-training and target datasets, for each relation label names in meta-training datasets, we discard it if it overlaps with a relation label name in target RE datasets (i.e, two identical phrases appear in two names). 

\paragraph{Zero-shot settings.} We randomly select $m$ relations from FewRel and Wiki-ZSL as zero-shot relations~\cite{chen-li-2021-zs}. We repeat the experiment 5 times for random selection of $m$ relations, and report the average results. We also vary $m$ to examine how performance is affected. We use Precision (P), Recall (R), and Macro-F1 as the evaluation metrics for RC. For RTE, evaluating single triplet extraction involves only one possible triplet for each sentence, hence the metric used is Accuracy (Acc.)~\cite{chia2022relationprompt}. Note that the randomly sampled zero-shot relations may share similar semantics with some relations appearing in the meta-training datasets. However, since the relation schemas of meta-training and target datasets are different, we can still evaluate the zero-shot transfer learning ability of the models. In other words, the models are expected to understand the zero-shot relation semantics solely based on relation names.

\paragraph{Few-shot settings.} We conduct experiments on the public benchmark dataset FewRel~\cite{han-etal-2018-fewrel}, which releases 80 relations and each relation owns 700 triple instances in total. For RC, following the standard configuration of FewRel~\cite{han-etal-2018-fewrel}, we conducted experiments in these settings: 5-way-1-shot, 5-way-5-shot, 10-way-1-shot and 10-way-5-shot. For RTE, we follow previous work~\cite{yu2020bridging} to adopt the 5-way-5-shot and 10-way-10-shot settings. Concretely, a relational triple is correct if and only if the spans of the head and tail entity are correctly identified and the associated relation is also predicted correctly. We adopt the standard Micro F1 score to evaluate the results and report the averages over 5 randomly initialized runs. Because the maximum length limitation of LLMs restricts to put too many in-context examples at once, we concatenate the maximum number of training examples satisfying the input length and discard the rest examples specially in 10-way-10-shot setting. And we should point out that this is one of the limitations of \textsc{Micre}. More generally, in-context leaning paradigm is suitable for very few examples, making it difficult to better utilize more training examples compared to traditional methods. 

\paragraph{Baselines.} We consider both supervised fine-tuned methods and zero/few-shot prompting methods. For \textbf{Zero-shot RC}, we make comparisons with state-of-the-art matching-based methods ESIM~\cite{levy-etal-2017-zero}, ZS-BERT~\cite{chen-li-2021-zs}, PromptMatch~\cite{sainz2021label} and RE-Matching~\cite{zhao-etal-2023-matching}. We also compare a seq2seq-based method RelationPrompt~\cite{chia2022relationprompt} and two zero-shot prompting methods Vanilla and SumAsk~\cite{li2023revisiting} with GPT-3.5~\cite{openai2022}. For \textbf{Zero-shot RTE}, we provide three baseline methods for comparison: TableSequence~\cite{wang2020two}, RelationPrompt~\cite{chia2022relationprompt} and ZETT~\cite{kim2023zero}. For \textbf{Few-shot RC}, we make comparisons with the following state-of-the-art baselines including traditional few-shot learning methods ProtoNet~\cite{snell2017prototypical} and MAML~\cite{finn2017model}, besides the pre-training enhanced methods CP~\cite{peng2020learning}, HCRP~\cite{han2021exploring}, LPD~\cite{zhang2022better}, HDN~\cite{zhang-etal-2023-hypernetwork} and DeepStruct~\cite{wang2022deepstruct}. We also compare a recent promising chain of thought and few-shot prompting method CoT-ER~\cite{ma2023chain}. For \textbf{Few-shot RTE}, we select supervised learning methods FT-BERT~\cite{devlin2019bert}, FastRE~\cite{li2022fastre} and CasRel~\cite{wei2020novel}, besides the few-shot learning methods MPE~\cite{yu2020bridging}, StructShot~\cite{yang2020simple}, PA-CRF~\cite{cong2021few}, RelATE~\cite{cong2022relation} and MG-FTE~\cite{yang2023mutually}.

\begin{table*}[t]
\small
\centering\setlength{\tabcolsep}{0.3mm}
\begin{tabular}{cccccccccccccccccccc}
\toprule
\multirow{3}{*}{\textbf{Method}} & \multicolumn{9}{c}{\textbf{Wiki-ZSL}} & \multicolumn{9}{c}{\textbf{FewRel}} & \multirow{3}{*}{\textbf{Avg.}}\\
& \multicolumn{3}{c}{m=5} & \multicolumn{3}{c}{m=10} & \multicolumn{3}{c}{m=15} & \multicolumn{3}{c}{m=5} & \multicolumn{3}{c}{m=10} & \multicolumn{3}{c}{m=15} & \\
\cmidrule{2-19}
& {P} & {R} & {F1} & {P} & {R} & {F1} & {P} & {R} & {F1} & {P} & {R} & {F1} & {P} & {R} & {F1} & {P} & {R} & {F1} & \\
\midrule
{ESIM} & {48.58} & {47.74} & {48.16} & {44.12} & {45.46} & {44.78} & {27.31} & {29.62} & {28.42} & {56.27} & {58.44} & {57.33} & {42.89} & {44.17} & {43.52} & {29.15} & {31.59} & {30.32} & {42.09} \\
{ZS-BERT} & {71.54} & {72.39} & {71.96} & {60.51} & {60.98} & {60.74} & {34.12} & {34.38} & {34.25} & {76.96} & {78.86} & {77.90} & {56.92} & {57.59} & {57.25} & {35.54} & {38.19} & {36.82} & {56.49} \\
{PromptMatch} & \underline{77.39} & {75.90} & {76.63} & {71.86} & {71.14} & {71.50} & {62.13} & {61.76} & {61.95} & \underline{91.14} & {90.86} & \underline{91.00} & \underline{83.05} & \underline{82.55} & \underline{82.80} & {72.83} & {72.10} & {72.46} & {76.06} \\
{RelationPrompt} & {70.66} & {83.75} & {76.63} & {68.51} & \underline{74.76} & {71.50} & {63.69} & \underline{67.93} & {65.74} & {90.15} & {88.50} & {89.30} & {80.33} & {79.62} & {79.96} & \textbf{74.33} & {72.51} & {73.40} & {76.09} \\
{RE-Matching} & \textbf{78.19} & \underline{78.41} & \textbf{78.30} & \textbf{74.39} & {73.54} & \textbf{73.96} & \textbf{67.31} & {67.33} & \underline{67.32} & \textbf{92.82} & \textbf{92.34} & \textbf{92.58} & \textbf{83.21} & \textbf{82.64} & \textbf{82.93} & \underline{73.80} & \underline{73.52} & \underline{73.66} & \textbf{78.13} \\
{Vanilla w/ GPT-3.5} & {64.47} & {70.83} & {67.50} & {41.83} & {46.22} & {43.92} & {23.17} & {27.82} & {25.28} & {67.41} & {72.97} & {70.08} & {42.48} & {46.26} & {44.29} & {25.71} & {27.77} & {26.70} & {46.30} \\
{SumAsk w/ GPT-3.5} & {75.64} & {70.96} & {73.23} & {62.31} & {61.08} & {61.69} & {43.55} & {40.27} & {41.85} & {78.27} & {72.55} & {75.30} & {64.77} & {60.94} & {62.80} & {44.76} & {41.13} & {42.87} & {59.62} \\
\midrule
{\textsc{Micre} w/ GPT-2} & {44.63} & {48.63} & {46.54} & {37.18} & {37.88} & {37.53} & {25.74} & {26.62} & {26.17} & {50.52} & {54.94} & {52.64} & {41.83} & {43.24} & {42.52} & {32.64} & {33.52} & {33.07} & {39.75} \\
{\textsc{Micre} w/ GPT-2-large} & {68.22} & {70.50} & {69.34} & {64.78} & {65.62} & {65.20} & {54.51} & {55.87} & {55.18} & {80.64} & {83.52} & {82.05} & {68.41} & {69.99} & {69.19} & {60.55} & {62.68} & {61.60} & {67.09} \\
{\textsc{Micre} w/ GPT-2-XL} & {70.75} & {73.66} & {72.18} & {67.43} & {69.75} & {68.57} & {62.30} & {63.64} & {62.96} & {86.74} & {89.23} & {87.97} & {77.85} & {78.73} & {78.29} & {68.60} & {69.52} & {69.06} & {73.17} \\
{\textsc{Micre} w/ T5-base} & {56.53} & {58.66} & {57.58} & {45.73} & {47.28} & {46.49} & {26.74} & {27.31} & {27.02} & {73.24} & {76.54} & {74.85} & {65.10} & {66.76} & {65.92} & {43.62} & {44.60} & {44.10} & {52.66} \\
{\textsc{Micre} w/ T5-large} & {70.25} & {72.55} & {71.38} & {65.37} & {67.44} & {66.39} & {57.45} & {58.39} & {57.92} & {82.43} & {84.63} & {83.52} & {71.72} & {73.51} & {72.60} & {60.57} & {61.98} & {61.27} & {68.85} \\
{\textsc{Micre} w/ T5-3B} & {74.75} & {77.36} & {76.03} & {70.64} & {71.89} & {71.30} & {64.11} & {65.43} & {64.76} & {88.23} & {89.77} & {88.99} & {78.82} & {80.53} & {79.67} & {70.28} & {71.69} & {70.98} & {75.29} \\
{\textsc{Micre} w/ LLaMA} & {76.46} & \textbf{78.53} & \underline{77.48} & \underline{72.36} & \textbf{74.88} & \underline{73.60} & \underline{67.14} & \textbf{68.87} & \textbf{67.99} & {89.34} & \underline{91.88} & {90.59} & {80.67} & {82.31} & {81.48} & {73.74} & \textbf{75.83} & \textbf{74.77} & \underline{77.65} \\
\bottomrule
\end{tabular}
\caption{Zero-shot RC results on Wiki-ZSL and FewRel datasets. Best results are in \textbf{bold} and the second best results are marked with \underline{underline}. \textbf{Avg.} denotes the average of all the Macro-F1 scores in six settings.}
\label{zerorc}
\end{table*}

\paragraph{Experiment Details.} 
For meta-training, we use a batch size of 4, learning rate of 1e-4 and a block size of 512, training the model for 100,000 max steps with 16-shot learning. To save memory during meta-training, we use deepspeed~\cite{rasley2020deepspeed} and adopt parameter efficient tuning technique LoRA~\cite{hu2021lora} for model training with the rank $r$ to 8 and the merging ratio $\alpha$ to 32. As for the base LLMs, we use popular open-source models such as GPT-2~\cite{radford2019language}, T5~\cite{raffel2020exploring} and LLaMA~\cite{touvron2023llama}. Specifically, we adopt GPT-2 (117M), GPT-2-large (770M), GPT-2-XL (1.5B), T5-base (220M), T5-large (770M), T5-3B and LLaMA-7B for experiments. We should note that these LLMs without fine-tuning cannot perform ICL in RE. We empirically discover that they are unable to understand the structural sentences in ICL paradigm and recover the relation labels of test examples in low-shot settings. 

\subsection{Main Results}
\paragraph{Zero-shot RC results.} The main results of zero-shot RC are summarized in Table~\ref{zerorc}, where LLMs with meta in-context training achieve competitive results compared to supervised fine-tuned RC methods and zero-shot prompting RC methods over two datasests when varying numbers of unseen relations. We have two findings about the model parameter scales and overall performances. First, the larger the model scale, the more notable the enhancement in performance. With small model scale, \textsc{Micre} with GPT-2 even underperforms ESIM. Second, encoder-decoder models seem to achieve better ICL performance than decoder-only models with similar model scales (eg., GPT-2-large and T5-large), which is due to the positive role of encoders in language understanding tasks. For the zero-shot RC task, as $m$ increases, it is straightforward that models are difficult to predict the right relation since the possible choices have increased. Notably, the proposed \textsc{Micre} with LLaMA delivers superior results compared to state-of-the-art method RE-Matching when dealing with more unseen relations. Such results not only validate the effectiveness of meta in-context training, but indicate \textsc{Micre} is less sensitive to the number of relations compared to baselines. Another finding is that the recall of \textsc{Micre} with LLaMA achieves the best or second best results in 5 out of 6 settings, which may be related to our zero-shot prompting strategy. Because we enumerate each relation and subject to prompt \textsc{Micre} to generate its corresponding object, ensuring the final recall but slightly harming the final precision.

\begin{table}[t]
\small
\centering\setlength{\tabcolsep}{0.3mm}
\begin{tabular}{cccccccc}
\toprule
\multirow{3}{*}{\textbf{Method}} & \multicolumn{3}{c}{\textbf{Wiki-ZSL}} & \multicolumn{3}{c}{\textbf{FewRel}} & \multirow{3}{*}{\textbf{Avg.}} \\
& {m=5} & {m=10} & {m=15} & {m=5} & {m=10} & {m=15} & \\
\midrule
{TableSequence} & {14.47} & {9.61} & {9.20} & {11.82} & {12.54} & {11.65} & {11.55} \\
{RelationPrompt} & {16.74} & {12.13} & {10.47} & {24.36} & {21.45} & {20.24} & {17.57} \\
{ZETT} & {21.49} & {17.27} & {12.78} & {30.71} & {27.90} & {26.17} & {22.72} \\
\midrule
{\textsc{Micre} w/ GPT-2} & {14.77} & {10.62} & {6.90} & {12.88} & {9.20} & {6.93} & {10.22} \\
{\textsc{Micre} w/ GPT-2-large} & {19.50} & {17.73} & {14.03} & {28.36} & {24.75} & {17.66} & {20.34} \\
{\textsc{Micre} w/ GPT-2-XL} & {21.56} & {19.61} & {15.75} & {31.97} & {28.44} & {21.20} & {23.09} \\
{\textsc{Micre} w/ T5-base} & {17.43} & {15.35} & {11.66} & {27.34} & {23.54} & {15.98} & {18.55} \\
{\textsc{Micre} w/ T5-large} & {20.64} & {17.87} & {14.53} & {29.58} & {26.40} & {18.05} & {21.18} \\
{\textsc{Micre} w/ T5-3B} & \underline{25.20} & \underline{23.65} & \underline{21.80} & \underline{36.75} & \underline{33.18} & \underline{30.44} & \underline{28.50} \\
{\textsc{Micre} w/ LLaMA} & \textbf{27.74} & \textbf{24.64} & \textbf{22.23} & \textbf{37.53} & \textbf{34.77} & \textbf{32.42} & \textbf{29.89} \\
\bottomrule
\end{tabular}
\caption{Zero-shot RTE results on Wiki-ZSL and FewRel datasets. Best results are in \textbf{bold} and the second best results are marked with \underline{underline}. \textbf{Avg.} denotes the average of all the accuracy scores.}
\label{zerorte}
\end{table}

\begin{table*}[t]
\small
\centering\setlength{\tabcolsep}{0.5mm}
\begin{tabular}{c|c|c|c|c|c||c|c|c|c}
\toprule
{\textbf{Method}} & \textbf{5-way-1-shot} & \textbf{5-way-5-shot} & \textbf{10-way-1-shot} & \textbf{10-way-5-shot} & \textbf{Avg.} & {\textbf{Method}} & \textbf{5-way-5-shot} & \textbf{10-way-10-shot} & \textbf{Avg.} \\
\midrule
{ProtoNet} & {82.92} & {91.32} & {73.24} & {83.68} & {82.79} & {FT-BERT} & {4.71} & {2.94} & {3.83} \\
{MAML} & {82.93} & {86.21} & {73.20} & {76.06} & {79.60} & {FastRE} & {7.73} & {6.82} & {7.28} \\
{CP} & {88.29} & {92.77} & {80.50} & {88.61} & {87.54} & {CasRel} & {2.11} & {2.04} & {2.08} \\
{HCRP} & {90.89} & {92.90} & {83.17} & {86.43} & {88.35} & {MPE} & {23.34} & {12.08} & {17.71} \\
{LPD} & {93.51} & {94.33} & {87.77} & {89.19} & {91.20} & {StructShot} & {25.94} & {20.28} & {23.11} \\
{HDN} & {95.46} & {96.59} & {89.34} & {92.46} & {93.46} & {PA-CRF} & {34.14} & {30.44} & {32.29} \\
{DeepStruct} & \textbf{98.40} & \textbf{100.00} & \textbf{97.80} & \textbf{99.80} & \textbf{99.00} & {RelATE} & {42.32} & {40.93} & {41.63} \\
{CoT-ER w/ GPT-3} & \underline{97.40} & {97.00} & {92.10} & \underline{94.70} & \underline{95.30} & {MG-FTE} & {55.17} & {53.33} & {54.25} \\
\midrule
{\textsc{Micre} w/ GPT-2} & {71.53} & {72.33} & {65.84} & {66.15} & {68.96} & {-} & {43.44} & {40.64} & {42.04} \\
{\textsc{Micre} w/ GPT-large} & {85.37} & {85.98} & {78.34} & {80.46} & {82.54} & {-} & {46.36} & {44.20} & {45.28} \\
{\textsc{Micre} w/ GPT-XL} & {89.38} & {90.58} & {82.10} & {83.74} & {87.45} & {-} & {52.66} & {50.82} & {51.74} \\
{\textsc{Micre} w/ T5-base} & {76.24} & {78.92} & {72.75} & {73.55} & {75.37} & {-} & {45.08} & {44.00} & {44.54} \\
{\textsc{Micre} w/ T5-large} & {87.77} & {88.46} & {79.93} & {82.64} & {84.70} & {-} & {48.64} & {47.59} & {48.12} \\
{\textsc{Micre} w/ T5-3B} & {93.77} & {94.30} & {88.45} & {89.56} & {91.52} & {-} & \underline{56.35} & \underline{53.57} & \underline{54.96} \\
{\textsc{Micre} w/ LLaMA} & {95.74} & \underline{97.11} & \underline{93.36} & {94.25} & {95.12} & {-} & \textbf{59.82} & \textbf{56.75} & \textbf{58.29} \\
\bottomrule
\end{tabular}
\caption{Few-shot RC results (left) and few-shot RTE results (right) on FewRel. Best results are in \textbf{bold} and the second best results are marked with \underline{underline}. \textbf{Avg.} denotes the average of all the Micro-F1 scores.}
\label{fewrc}
\end{table*}

\paragraph{Zero-shot RTE results.} The main results of zero-shot RTE by varying $m$ unseen relations on FewRel and Wiki-ZSL are summarized in Table~\ref{zerorte}, where \textsc{Micre} with foundation models that have more than 1B parameters consistently outperforms existing methods across different settings. LLaMA achieves up to 9.45 and 6.87 higher accuracy than the existing state-of-the-art model, ZETT on Wiki-ZSL and FewRel datasets, respectively. Because extracting relational triples from texts is a challenging structure prediction task, the meta-training makes \textsc{Micre} recover the semantics of the RTE task during zero-shot inference, where the model output is very similar with the meta-training output. However, we note that with the same foundation models, \textsc{Micre} achieves less satisfied performance compared to existing baselines. Specifically, the average scores of \textsc{Micre} with GPT-2 and T5-base are 10.22 and 18.55, respectively, where RelationPrompt uses GPT-2 and ZETT uses T5-base but they all delivers better results. This indicates that achieving noticeable ICL results in LLMs requires the relatively large-scale model parameters. As the model scale increases, the advantages of \textsc{Micre} become more apparent, where T5-3B and LLaMA both show much better performances than previous methods.

\paragraph{Few-shot RC results.} The main results of few-shot RC are summarized in Table~\ref{fewrc} (left). We observe strong few-shot RC performance of \textsc{Micre} on FewRel. This suggests that the meta in-context training is beneficial in low-resource regimes via transferring knowledge from similar tasks. First, compared to few-shot and pre-training enhanced RC methods (e.g., LPD and HDN), \textsc{Micre} with LLaMA achieve more superior performances, and \textsc{Micre} with smaller base models can show competitive results. Second, compared to CoT-ER which is based on elaborated few-shot prompting and GPT-3, \textsc{Micre} with LLaMA basically delivers similar average scores, which indicates that meta in-context training successfully boosts the ICL abilities of open-source LLMs in RE. Third, \textsc{Micre} lags behind DeepStruct which is based on a pre-trained 10B parameter encoder-decoder language model GLM~\cite{du-etal-2022-glm} and is pre-trained on a collection of large-scale RE corpus. Despite its excellent performance with multi-task fine-tuning, DeepStruct produces unattractive zero-shot transferring ability after pre-training~\cite{wang2022deepstruct}. In contrast, \textsc{Micre} appears to be able to adapt to new data, presenting a fair and strong performance with in-context learning. Another explanation is that in-context training examples help the model better understand the new RE tasks, such as the concrete output format of each task. Finally, \textsc{Micre} with GPT-2 and T5-base still cannot surpass classic few-shot methods ProtoNet and MAML. Compared to pre-train then fine-tune paradigm, showing in-context learning ability tends to require large-scale model parameters. As the exceptional performance of 10B GLM based DeepStruct, meta-training on larger base models may bring better in-context learning results and is worth exploring in the future.

\paragraph{Few-shot RTE results.} Table~\ref{fewrc} (right) reports the few-shot RTE results of \textsc{Micre} against other baseline models on FewRel. It can be seen that, overall, \textsc{Micre} with LLMs significantly outperforms all competitive methods and achieves new state-of-the-art in two few-shot settings, which highlights the pivot role of meta in-context training. 
Notably, even with GPT-2 and T5-base, \textsc{Micre} still outperforms most strong baselines such as PA-CRF and RelATE. Similar with zero-shot RTE results, the few-shot RTE results seem to prove that generative methods are more effective in handling the low-shot RTE task. Moreover, current few-shot RTE methods first meta-train on the subset of entire dataset and then are evaluated on the test set. Thus the training and testing data are in the similar distribution and same domain. But \textsc{Micre} showcases notable distribution and domain adaptation ability.

\subsection{Discussions}
\begin{figure}[t]
\includegraphics[width=0.23\textwidth]{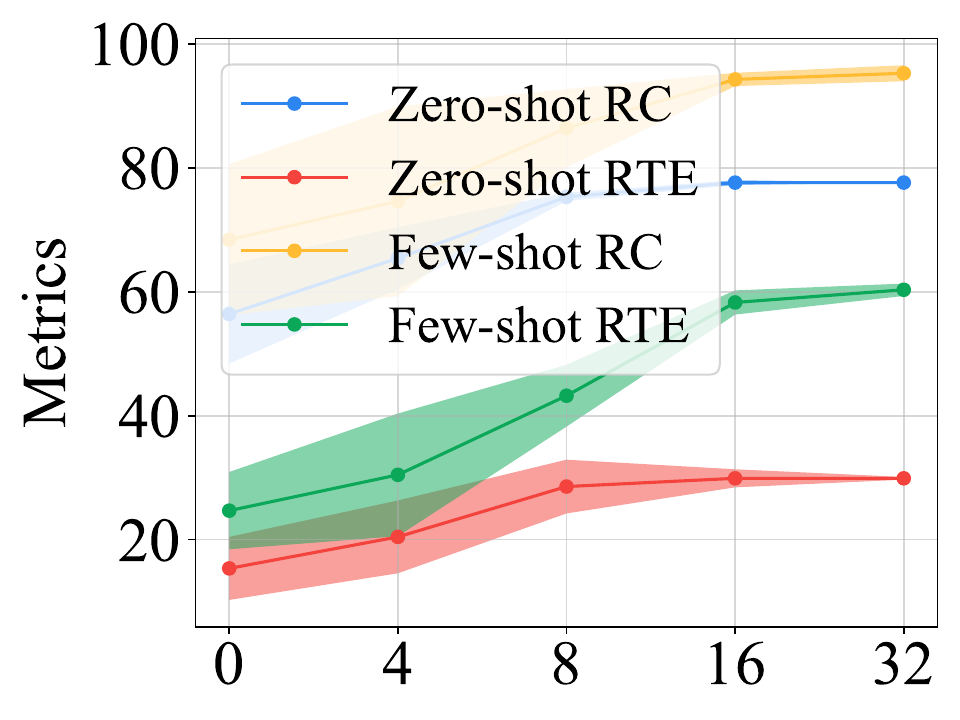} \label{shot} \hspace{1mm}
\includegraphics[width=0.23\textwidth]{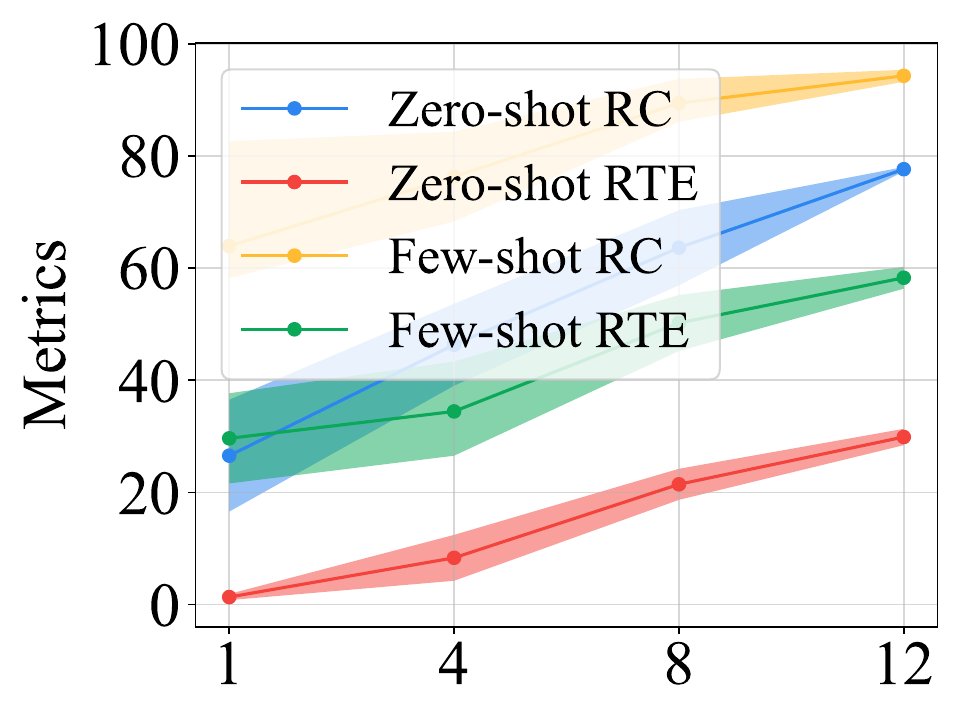} \label{task}
\caption{Ablation on the number of training examples $k$ (left) and meta-training datasets $C$ (right) in four settings. The metric of each setting corresponds to its \textbf{Avg.} score.}
\label{number}
\end{figure}

\paragraph{Number of in-context training examples.} We vary the number of training examples $k$ from 0, 4, 8, 16 to 32. In-context learning with $k$ = 0 is equivalent to the zero-shot method. Results are shown in Figure~\ref{number}. Generally, increasing $k$ helps across all tasks and settings. Besides, the few-shot setting seems to have higher variance than zero-shot setting. And increasing $k$ consistently reduces the average performance variance. Especially, for zero-shot cases, performance tends to stabilize when $k$ is greater than 8. This indicates that zero-shot prompting is less sensitive to the number of in-context training examples during meta-training as no training example is provided at inference time. In contrast, meta-training with more examples brings significant improvements for few-shot learning. However, we additionally find that the performance tends to saturate when $k$ is closer to 16~\cite{min2022metaicl}. The saturate phenomenon is likely because the sequence length limit of the language model makes it hard to encode many training examples, which is one of the limitations of the attention technique~\cite{vaswani2017attention}. 

\paragraph{Number of meta-training datasets.} To see the impact of the number of meta-training datasets, we subsample 1, 4, 8 meta-training datasets out of 12 in four experimental settings. For each, we use three different random seeds to additionally see the impact of the choice of meta-training datasets. Figure~\ref{number} shows the results. On average, low-shot performances generally increase as the number of datasets increase, which is consistent with results in previous work~\cite{mishra2022cross,wei2022finetuned,min2022metaicl}. Nonetheless, different choices of meta-training datasets brings nonnegligible variance, indicating that a choice of meta-training gives substantial impact in performance. This can be attributed to multiple reasons. On the one hand, the varying amount of training data in different datasets leads to the training effectiveness of the model. On the other hand, the data domains and distributions between meta-training datasets and target datasets also play a key role in model performances.

\begin{table}[t]
\small
\centering\setlength{\tabcolsep}{2mm}
\begin{tabular}{c|c|cc|cc}
\toprule
\multirow{2}{*}{\textbf{Train labels}} & \multirow{2}{*}{\textbf{Test labels}} & \multicolumn{2}{c}{\textbf{Zero-shot}} & \multicolumn{2}{c}{\textbf{Few-shot}} \\
& & {RC} & {RTE} & {RC} & {RTE} \\
\midrule
{-} & {Original} & {0.00} & {0.00} & {0.00} & {0.00} \\
{Original} & {Original} & {77.65} & {29.89} & {94.31} & {58.29} \\
{Original} & {Replaced} & {-} & {-} & {68.64} & {22.11} \\
{Replaced} & {Original} & {28.62} & {1.74} & {84.45} & {45.44} \\
{Replaced} & {Replaced} & {-} & {-} & {58.43} & {15.38} \\
\bottomrule
\end{tabular}
\caption{Ablation about semantic hints of similar relations. Original and Replaced indicate original label words and labels that are replaced to special tokens, respectively. The first row denotes the LLaMA performance without meta-training. We report \textbf{Avg.} scores.}
\label{hint}
\end{table}

\paragraph{Semantic hints of similar relations.} Although we ensure that the data distributions and relation schemas during the meta-training and inference phases are different, in fact, there are inevitably similar semantic relations between meta-training and inference relation sets. We use relation label words taken from the original datasets, which contain semantic hints that express what each relation label is supposed to mean. If the model is truly learning the relation in-context, it should generalize when label words are replaced with other English words, e.g., \textit{date\_of\_birth} is replaced with token $R1$, thus not giving any hints about the relation semantics. To this end, we substitute each relation label in meta-training with $Ri, i \in [1, I]$ where $I$ is the total number of relations in meta-training datasets. At inference time, we also perform similar operations to evaluate the in-context relation learning ability. The results are summarized in Table~\ref{hint}. Note that when test labels are replaced, \textsc{Micre} is unable to perform zero-shot RC and RTE using tabular prompting. First, with test labels be replaced, the overall results suffer grave declines. This indicates that having semantic hints from relation label words is a necessary condition for LLMs to perform low-shot RE tasks. Compared to original LLaMA, meta-training on replaced training labels consistently delivers considerable improvements in few-shot RC and RTE, where \textsc{Micre} actually benefits from training on the replaced data and improves its in-context learning ability on new RE task. Still, overall performance is relatively poor compared to training on original labels, which implies that learning from relation label words helps the model better capture semantic differences between various relations. And the model can utilize the relation semantic knowledge during inference on target RE datasets.

\begin{figure}[t]
	\centering
	\includegraphics[width=\linewidth]{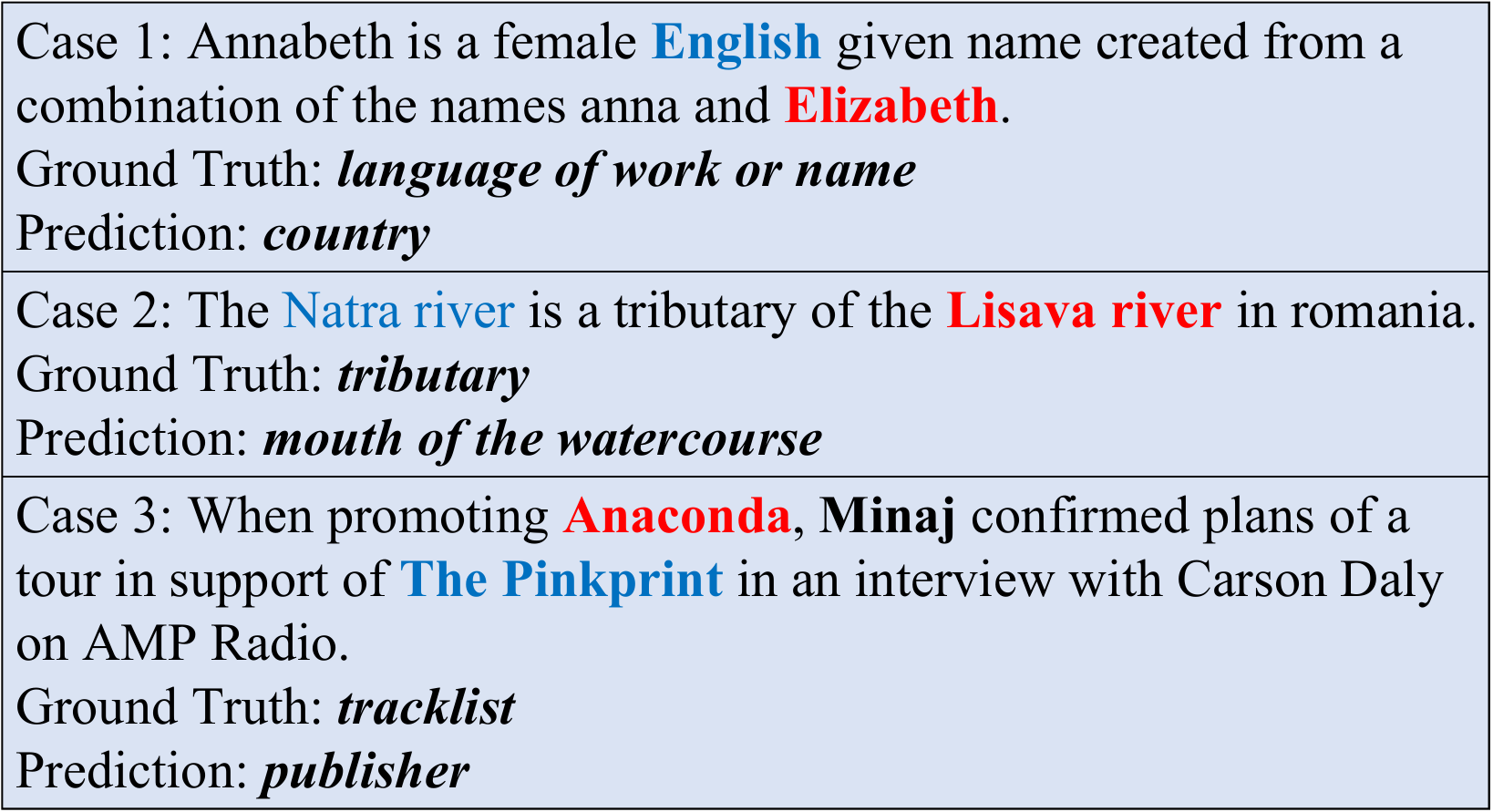}
	\caption{List of three cases. The entities are highlighted in color.}
	\label{case}
\end{figure}

\paragraph{Error analysis.} We categorize three types of incorrectly predicted unseen relations for analysis and provide an example illustrated in Figure~\ref{case}. (1) The true relation is not appropriate because it comes from distant supervision. It shows the noise originated from distant labeling. That is, we cannot identify the relation between \textit{Elizabeth} and \textit{English} is \textit{language of work or name} in this specific sentence. They just happened to appear together and their relation recorded in Wikidata is \textit{language of work or name}. (2) The predicted relation is ambiguous because it is hard to identify the order of subject and object. The golden relation and predicted relation have very similar semantics because they are reciprocal in FewRel. Unfortunately, \textsc{Micre} frequently reverses the subject and object corresponding to these two relations because it treats the relational triple (\textit{Lisava river, tributary, Natra river}) as (\textit{Lisava river, tributary of, Natra river}). This indicates that relying solely on relation label names may bring ambiguity. (3) The predicted relation is not precise for the targeted entity pair but may be suitable for other entities that also appear in the sentence. The targeted entities are \textit{Anaconda} and \textit{The Pinkprint}, and \textsc{Micre} yields \textit{publisher} as the prediction, which is actually correct if the targeted entities are \textit{Anaconda} and \textit{Minaj}. This shows \textsc{Micre} is able to infer the possible relation for entities in the given sentence. When we prompt the model with relation \textit{publisher}, the model output the subject \textit{Anaconda} and object \textit{Minaj} with the max probability. As they are all valid spans in original sentence, we consider \textit{publisher} as the true relation. This also hinders the capability of \textsc{Micre} in extracting overlapping relational triples. More general methods are worth exploring in the future.

\section{Conclusion}
In this work, we introduce \textsc{Micre}, a new zero and few-shot learning method where an LLM is meta-trained to learn to in-context learn relations, i.e. condition on training examples to recover the new relation semantics and make predictions. \textsc{Micre} outperforms a range of strong baselines including supervised fine-tuning and in-context learning without meta-training methods. Besides, it achieves competitive results compared to current state-of-the-art task-specific models. We also analyze the advantages and limitations of \textsc{Micre}, encouraging more effective methods in the future research.

\section*{Acknowledgments}
We thank the reviewers for their insightful comments. This work was supported by National Science Foundation of China (Grant Nos.62376057) and the Start-up Research Fund of Southeast University (RF1028623234). All opinions are of the authors and do not reflect the view of sponsors.

\bibliographystyle{named}
\bibliography{ijcai24}

\end{document}